\documentclass[Conference]{IEEEtran}
\usepackage{multirow}
\usepackage{multicol}
\usepackage{lipsum}
\usepackage{graphicx}
\usepackage{graphics}
\usepackage{amsmath}
\usepackage{amsbsy}
\usepackage{blindtext}
\usepackage{tcolorbox}
\usepackage{amssymb}
\usepackage{scalerel}
\usepackage{mathabx}
\usepackage{verbatim}
\usepackage{booktabs}
\usepackage{rotating,tabularx}
\usepackage{mathrsfs} 
\usepackage{hyperref}

\usepackage{adjustbox}
\usepackage{soul}
\usepackage{xcolor}
\usepackage{cite}
\usepackage{tikz}
\usepackage{color, colortbl}
\usepackage[first=0,last=9]{lcg}
\definecolor{LightGray}{rgb}{0.7,0.7,0.7}

\usepackage[wby]{callouts}
\usetikzlibrary{shapes.multipart}

\usepackage{array}
\usepackage{makecell}

\allowdisplaybreaks

\usepackage{amsthm}
\theoremstyle{definition}

\theoremstyle{remark}

\usepackage[utf8]{inputenc}
\usepackage[english]{babel}

\usepackage[caption=false,font=footnotesize]{subfig}
\usepackage[T1]{fontenc}
\usepackage{scalerel,stackengine}
\newcommand\reallywidecheck[1]{%
\savestack{\tmpbox}{\stretchto{%
  \scaleto{%
    \scalerel*[\widthof{\ensuremath{#1}}]{\kern-.6pt\bigwedge\kern-.6pt}%
    {\rule[-\textheight/2]{1ex}{\textheight}}%WIDTH-LIMITED BIG WEDGE
  }{\textheight}% 
}{0.5ex}}%
\stackon[1pt]{#1}{\scalebox{-1}{\tmpbox}}%
}
\stackMath
\IEEEoverridecommandlockouts
\IEEEoverridecommandlockouts

\newcommand*{\rn}{\textcolor{black}}

\newcommand*{\ra}{\textcolor{black}}

\usepackage{bm} 

\usepackage[margin=1in,footskip=0.4in]{geometry}

\newif\ifarxiv
\arxivtrue
\ifarxiv
\renewcommand{\baselinestretch}{.9635}
\IEEEtriggeratref{40}
\else
\renewcommand{\baselinestretch}{.910}
 \IEEEtriggeratref{40}
\fi

\begin{document}

\title{\LARGE\bf
\rn{Attention-Enhanced Graph Filtering for False Data Injection Attack Detection and Localization}}

\author{\rn{Ruslan Abdulin,$^{\ast}$ Mohammad Rasoul Narimani$^{\ast}$
\thanks{\rn{${*}$: Department of Electrical and Computer Engineering, California State University, Northridge (CSUN). Rasoul.narimani@csun.edu. Support from NSF contract \#2523881
and \#2308498.}% 
}}}

\maketitle

\begin{abstract}

\ra{The increasing deployment of Internet-of-Things (IoT)-enabled measurement devices in modern power systems has expanded the cyberattack surface of the grid.
As a result, this critical infrastructure is increasingly exposed to cyberattacks, including false data injection attacks (FDIAs) that compromise measurement integrity and threaten reliable system operation. Existing FDIA detection methods primarily exploit spatial correlations and network topology using graph-based learning; however, these approaches often rely on high-dimensional representations and shallow classifiers, limiting their ability to capture local structural dependencies and global contextual relationships. Moreover, naively incorporating Transformer architectures can result in overly deep models that struggle to model localized grid dynamics.
This paper proposes a joint FDIA detection and localization framework that integrates auto-regressive moving average (ARMA) graph convolutional filters with an Encoder-Only Transformer architecture. The ARMA-based graph filters provide robust, topology-aware feature extraction and adaptability to abrupt spectral changes, while the Transformer encoder leverages self-attention to capture long-range dependencies among grid elements without sacrificing essential local context. The proposed method is evaluated using real-world load data from the New York Independent System Operator (NYISO) applied to the IEEE 14- and 300-bus systems. Numerical results demonstrate that the proposed model effectively exploits both the state and topology of the power grid, achieving high accuracy in detecting FDIA events and localizing compromised nodes.}
\end{abstract}
%\vspace{-0.25cm}

\section{Introduction}
\label{Introduction}

\ra{Modern power grids, often referred to as smart grids, rely on networked communication among their components to improve the efficiency of electricity generation, transmission, and distribution. Measurements collected by Phasor Measurement Units (PMUs) and Remote Terminal Units (RTUs) are transmitted to Supervisory Control and Data Acquisition (SCADA) systems and subsequently processed by power system state estimation (PSSE) algorithms to infer the system’s operating state. The output of PSSE is then utilized by the Energy Management System (EMS) to support operational decision-making. As a result, the secure operation of power systems critically depends on the integrity and reliability of measurement data. An intruder can take advantage of this dependency by injecting false data into measurement streams, corrupting the state estimation process and misleading grid operators about the true system state~\cite{boyaci2021joint}.}

\ra{False data injection attacks (FDIAs) pose a significant threat because they can be designed to bypass traditional model-based detection schemes, such as bad data detection (BDD) methods based on the largest normalized residual test (LNRT). Prior studies have demonstrated that carefully crafted false measurements satisfying the power flow equations can evade these protection mechanisms when they rely solely on system state consistency \cite{deng2016false, bi2011defending, wang2023valid}. Consequently, if a power grid is protected only by conventional detection methods, a sufficiently informed adversary may successfully conduct stealthy attacks. Moreover, while detection techniques can indicate the presence of an attack, they do not identify the compromised locations, rendering the network unreliable for corrective control actions. This limitation motivates the need for joint FDIA detection and localization methods to ensure practical and effective mitigation of cyberattacks in power systems.}

\ra{Beyond model-based techniques, data-driven approaches have been widely investigated for FDIA detection and localization due to their ability to learn attack patterns directly from historical measurements without requiring detailed system parameters. When properly trained, such methods can adapt to diverse operating conditions and FDIA strategies. In recent years, deep learning models, including fully connected neural networks (FCNs), recurrent neural networks (RNNs), and convolutional neural networks (CNNs), have been proposed for power system cyberattack detection, driven by the increasing adoption of machine learning across engineering applications. Despite their expressive capabilities, these models can exhibit limited generalization when exposed to varying operating points and unseen attack types. Two primary limitations contribute to this challenge. First, conventional deep learning architectures do not explicitly account for the underlying network structure, limiting their ability to contextualize node behavior within the power grid topology. Second, they rely on rigid, input-invariant dependency patterns, which restrict their adaptability to heterogeneous and spatially varying FDIA scenarios.}

%Besides the model-based techniques, there are also data-driven detection and localization methods that do not depend on the system parameters as heavily. They require a set of historical data to learn how to generalize between healthy and attacked states. Such an approach lacks inherent in-depth insights about the particular power grid; nevertheless, if trained properly, data-driven algorithms demonstrate a strong ability to learn those specifics and adaptively guard the system against various FDIAs. In the recent past, deep learning models such as Fully-Connected Neural Network (FCN), Recurrent Neural Network (RNN), and Convolutional Neural Network (CNN) have been proposed for data-driven cyberattack detection due to the increasing popularity of the Machine Learning (ML) application across all fields. Despite their profound capabilities, these models show low generalization rates, causing them to fail frequently in identifying whether a power grid is under attack. There are two main reasons for that. First, these algorithms do not treat the network’s structure as equally important as the features, which limits their ability to contextualize the roles of nodes. Second, the conventional algorithms converge toward a rigid, input-invariant pattern. For example, CNN displays great results in multi-label image classification due to its ability to discern local features, but it applies the same learned filters at every spatial location. Similarly, an RNN is effective in sequence modeling, but the recurrence weights are the same for all time steps.

\ra{To overcome the lack of structural awareness, graph neural networks (GNNs) have been employed for FDIA detection by leveraging topology-aware message passing mechanisms that utilize the adjacency structure of power grids. Conventional GNN variants, such as graph convolutional networks (GCNs), demonstrate improved detection performance due to their ability to capture spatial dependencies; however, their reliance on first-order spectral approximations can lead to high memory and computational costs while limiting the effective receptive field. Chebyshev graph convolutional networks (CGCNs) address this limitation by employing higher-order polynomial spectral filters, enabling the aggregation of K-hop neighborhood information within a single layer \cite{defferrard2016convolutional}. When applied to large-scale power grids, CGCNs have achieved significantly higher FDIA detection performance than conventional machine learning models \cite{boyaci2022cyberattack}. Nevertheless, CGCNs are prone to over-smoothing and noise amplification at higher polynomial orders, which can blur subtle distinctions between attacked and healthy nodes and hinder precise localization \cite{cheng2024graph}. Graph neural networks with convolutional auto-regressive moving average filters (ARMAConv) mitigate these issues by employing learnable spectral filters that provide depth tolerance, enhanced noise robustness, and improved localization sensitivity \cite{bianchi2021graph}. Empirical evaluations on IEEE-57, IEEE-118, and IEEE-300 bus systems have shown that ARMAConv outperforms CGCN and other traditional approaches in FDIA localization tasks \cite{boyaci2021joint}. However, ARMAConv applies fixed aggregation patterns across samples, limiting its ability to adapt edge influences to input-specific conditions.}

\ra{To address the limitation of rigid dependency modeling, self-attention mechanisms have been introduced to enable adaptive, input-dependent relationship learning. The original Transformer architecture employs scaled dot-product attention to analyze information across multiple representation subspaces by dynamically assigning pairwise interaction weights among all elements of the input sequence \cite{vaswani2017attention}. In power grid applications, self-attention facilitates the modeling of non-local electrical correlations and simultaneous multi-region FDIA behaviors that may not be fully captured by localized spectral filtering alone. When combined with positional encoding, attention mechanisms can incorporate both relational and structural information, allowing routing patterns to depend on both the input data and the underlying grid structure. However, Transformer networks are computationally demanding and sensitive to noisy inputs, which limits the efficiency and robustness of their standalone application for large-scale power system cybersecurity tasks \cite{bagla2023noisy}.}

%To overcome the second limitation, the self-attention mechanism can be employed. Introduced in 2017, the Transformer network architecture operates the Scaled Dot-Product Attention algorithm to jointly analyze information across multiple lower-dimensional feature spaces at different positions\cite{vaswani2017attention}. Such a mechanism enables the model to dynamically compute the connectivity pattern by assigning pairwise interaction weights between all elements of the input. When combined with positional encoding to extract the position-dependent information, self-attention can determine its routing pattern based on both the input data and the underlying structure. In this way, the Transformer architecture addresses the fixed pattern issue. It is a significant factor in the context of detecting and localizing FDIAs, as there can be multiple types of attacks occurring across different areas of the power grid at the same time. On the other hand, the shortcomings of the Transformer network include high computational complexity and noise sensitivity, rendering the sole use of the model inefficient\cite{bagla2023noisy}.

\ra{In this work, we propose a joint FDIA detection and localization framework that integrates graph neural networks with convolutional auto-regressive moving average (ARMAConv) filters and an Encoder-Only Transformer architecture. The proposed model combines topology-aware spectral filtering for robust local feature extraction with self-attention–based global dependency modeling, enabling adaptive identification of attack patterns across the power grid. By excluding the Transformer decoder and leveraging ARMA-based smoothing and regularization, the framework achieves stable attention behavior while maintaining computational efficiency suitable for large-scale power system applications. The main contributions of this paper are summarized as:}

%In this work, we propose an FDIA detection and localization model that combines the GNN with convolutional auto-regressive moving average filters to enable structure-aware positional encodings and the Encoder-Transformer architecture to secure adaptive identification of the routing pattern. The resulting algorithm features stable self-attention, as well as reduced overfitting due to local smoothing/regularization from ARMAConv, and a low computational complexity trade-off resulting from noise reduction and the exclusion of the decoder section of the Transformer network.

\ra{
\begin{itemize}
\item A hybrid graph–attention framework (ACEOT) for joint FDIA detection and localization that captures both local structural dependencies and global contextual relationships in power grid data.
\item Positional encoding integrated with ARMA-based graph filtering provides persistent node identity, mitigating over-smoothing and noise amplification effects caused by malicious data injections and graph spectral filtering.
\item An attention-enhanced localization mechanism that enables adaptive, input-dependent modeling of inter-node dependencies, improving localization accuracy under diverse and spatially distributed FDIA scenarios.
\item Comprehensive validation on IEEE-14 and IEEE-300 bus systems using real-world NYISO load data, demonstrating improved detection and localization performance compared to state-of-the-art graph-based and deep learning benchmarks.
\end{itemize}}

\ra{The remainder of this paper is organized as follows. Section~\ref{Problem_formulation} formulates the FDIA detection and localization problem and introduces the associated constraints and graph-based representation. Section~\ref{sec: proposed_approach} presents the proposed ACEOT framework. Section~\ref{sec:experiments} describes the experimental setup and FDIA scenarios. Section~\ref{sec:res_disc} reports numerical results. Section~\ref{sec:Conclusion} concludes the paper.} \ifarxiv
\else
\ra{An extended version of this paper in~\cite{abdulin2026attention} reports the model-specific and general hyperparameter configurations obtained through systematic tuning for all baseline models and the proposed ACEOT framework on the IEEE-14 and IEEE-300 bus systems.} 
\fi

%The rest of this work is organized as follows. Section II describes the problem's theory, constraints, and graph-based representation. Section III presents and defines a novel approach for the joint detection and localization of FDIAs. Section IV introduces the experimental setup and lists the FDIA types utilized. Section V provides numerical results. Section VI concludes the work.

\section{Problem formulation}
\label{Problem_formulation}

\ra{This section formulates the FDIA detection and localization problem in power systems. We consider a graph-based representation of the power grid and define a physically consistent attack model in which adversarial measurement manipulations satisfy the full nonlinear AC power flow equations. The objective is to jointly detect the presence of an FDIA at the system level and localize compromised buses at the node level.}

\subsection{System Model and Threat Model}

\ra{A power network is modeled as a weighted, connected, undirected graph $\mathcal{G} = (\mathcal{V}, \mathcal{E}, w)$ where 
$\mathcal{V}$ denotes the set of buses, 
$\mathcal{E}$ represents transmission lines and transformers, and $w$ corresponds to a function $w : \mathcal{E} \rightarrow \mathbb{R}$ that assigns each edge a weight derived from network parameters. Each bus $i\in \mathcal{V}$ is associated with a system state consisting of voltage magnitude $V_i$ and phase angle $\theta_i$.} 
\ra{In modern power systems, measurements collected by RTUs and PMUs are transmitted to the SCADA system and processed by the PSSE module. PSSE computes an estimate of the system state vector $\mathbf{x} = [V_1, \theta_1, \dots, V_n, \theta_n]^T$ from a set of noisy measurements $\mathbf{z}$, which include active and reactive power injections and power flows.}

\ra{The state estimation problem is commonly formulated using weighted least squares estimation (WLSE) as
\begin{equation}
\hat{\mathbf{x}} = \arg\min_{\mathbf{x}} (\mathbf{z} - h(\mathbf{x}))^T \mathbf{R}^{-1} (\mathbf{z} - h(\mathbf{x})),
\end{equation}}

%\ra{where $h(\cdot)$ represents the nonlinear measurement function and $\mathbf{R}$ is the measurement error covariance matrix.}

where $h(\cdot)$ is the nonlinear measurement function and $\mathbf{R}$ the measurement noise covariance matrix.

\ra{The measurement vector $\mathbf{z}$ is composed of active and reactive power injections and line flows, given by the full AC power flow equations:
\begin{small}
\begin{align}
&\!\!\!\!P_i = \sum_{j \in \Omega_i} V_i V_j (G_{ij} \cos \theta_{ij} + B_{ij} \sin \theta_{ij}) = P_{Gi} -P_{Li}, \\
&\!\!\!\!Q_i = \sum_{j \in \Omega_i} V_i V_j (G_{ij} \sin \theta_{ij} - B_{ij} \cos \theta_{ij}) = Q_{Gi} - Q_{Li}, \\
&P_{ij} = V_i^2 (g_{si} + g_{ij}) - V_i V_j (g_{ij} \cos \theta_{ij} + b_{ij} \sin \theta_{ij}), \\
&Q_{ij} = -V_i^2 (b_{si} + b_{ij}) - V_i V_j (g_{ij} \sin \theta_{ij} - b_{ij} \cos \theta_{ij}),
\end{align}
\end{small}
where $G_{ij}$ and $B_{ij}$ are the conductance and susceptance matrices, respectively, and $\Omega_i$ denotes the set of buses connected to bus $i$.}

\ra{False data injection attacks aim to manipulate measurement data while preserving physical consistency. Let $\mathbf{z}_0$ denote the original measurement vector satisfying $\mathbf{z}_0 = h(\hat{\mathbf{x}})$. An FDIA constructs an attack vector $\mathbf{a}$ such that the compromised measurements
\begin{equation}
\mathbf{z}_a = \mathbf{z}_0 + \mathbf{a} = h(\tilde{\mathbf{x}}),
\end{equation}
correspond to an alternative system state $\tilde{\mathbf{x}}$. Because the injected data satisfy the full AC power flow equations, such attacks can bypass conventional bad data detection (BDD) schemes based solely on residual analysis.}

\subsection{Attack Constraints and Learning Formulation}

\ra{In practice, an adversary is constrained by limited access to measurement devices and by the difficulty of synchronizing attacks across geographically distant regions. As a result, realistic FDIAs typically affect localized areas of the network rather than the entire grid.}

\ra{Under these assumptions, we formulate FDIA detection and localization as a multi-label classification problem. Each bus $i \in \mathcal{V}$ is associated with a binary label $y_i \in \{0,1\}$, where $y_i = 1$ indicates that the bus is compromised by an FDIA and $y_i = 0$ denotes normal operation. In addition, a graph-level label $y_{\mathcal{G}} \in \{0,1\}$ represents the presence or absence of an FDIA in the network and is derived from the node-level labels by obtaining the maximum value among the predictions.}

\ra{Given a graph-structured input consisting of bus measurements and network topology, the learning objective is to jointly (i) detect whether the power grid is under attack and (ii) localize the compromised buses. This formulation enables practical mitigation by supporting both system-level awareness and targeted corrective actions.}

%\subsection{Theory}
%In general, an adversary’s goal when designing an FDIA is to maximize damage to the power network while minimizing the probability of being detected and localized. On the other hand, the goal of the BDD algorithm is to detect complex measurements $z$ collected from RTU and PMU that do not satisfy the system equations. 

%In power grids, the system state $x$ ($V_{i}$ and $\theta_{i}$ at each bus $i$) is computed with the use of the PSSE block, which iteratively reaches the estimation via weighted least squares estimation (WLSE)\ref{eq:system_state}, taking the complex measurements {z} as an input. Formally:

%\begin{equation}
%    \hat{x}=\min_{x}(z-h(x))^{T}R^{-1}(z-h(x))
%    \label{eq:system_state}
%\end{equation}
%Where R is the measurement's error covariance matrix and z is the set of complex measurements that include $P_{i}, Q_{i}, P_{ij}, Q_{ij}$ \ref{eq:pq}

%\begin{equation}
%    \begin{aligned}
    %P_{i}&=\sum_{j\in\Omega}V_{i}V_{j}(G_{ij}\cos{\theta_{ij}}+B_{ij}\sin{\theta_{ij}}) = P_{G_{i}}-P_{L_{i}} \\
    %Q_{i}&=\sum_{j\in\Omega}V_{i}V_{j}(G_{ij}\sin{\theta_{ij}}-B_{ij}\cos{\theta_{ij}}) = Q_{G_{i}}-Q_{L_{i}}  \\
   % P_{ij}&=V_{i}^{2}(g_{si}+g_{ij}) - V_{i}V_{j}(g_{ij}\cos{\theta_{ij}}+b_{ij}\sin{\theta_{ij}}) \\
   % Q_{ij}&=-V_{i}^{2}(b_{si}+b_{ij}) - V_{i}V_{j}(g_{ij}\sin{\theta_{ij}}-b_{ij}\cos{\theta_{ij}}) \\
    \label{eq:pq}

\section{Proposed approach}
\label{sec: proposed_approach}

\ra{This section presents the proposed joint FDIA detection and localization framework, referred to as the ARMAConv Encoder-Only Transformer (ACEOT). The framework is designed to address two key challenges in data-driven FDIA analysis: capturing topology-aware local dependencies while simultaneously modeling global, input-dependent interactions across the power grid. To this end, ACEOT integrates node-identity-aware positional encoding with graph convolutional auto-regressive moving average (ARMAConv) filters for robust local feature extraction, and an Encoder-Only Transformer architecture to enable adaptive modeling of long-range correlations. The individual components of the framework and their integration are described in the following subsections.}
%The proposed model combines a Positional Encoder, the ARMAConv algorithm, and the Transformer Encoder network into a robust solution. Below, we review each one of them separately and explain their roles in our proposed method: ARMAConv Encoder-Only Transformer (ACEOT).

\subsection{Positional-Encoder}

\ra{Positional encoding (PE) is employed to inject node-specific identity information into the model, enabling discrimination among buses with similar electrical characteristics. In power grids, neighboring buses may exhibit comparable active and reactive power measurements, which can lead to feature homogenization when graph-based spectral filtering is applied. Without positional cues, this effect can obscure subtle anomalies introduced by localized false data injection attacks.}

\ra{The original Transformer architectures employ deterministic sinusoidal encoding, which was originally proposed for sequence modeling tasks \cite{vaswani2017attention} and is still one of the most widely adopted baseline approaches. While effective in natural language processing, sinusoidal encodings assume an inherent ordering of elements along a linear axis \cite{takase2019positional}, an assumption that does not hold for power grid topologies where buses lack a natural sequential arrangement. Moreover, fixed encodings may limit adaptability when modeling complex, nonuniform structural relationships.}

\ra{To address these limitations, we adopt a learnable positional encoding implemented as an embedding layer, following recent advances in data-driven positional representations \cite{li2021learnable}. Specifically, each bus is assigned a trainable embedding vector that is optimized jointly with the model parameters via backpropagation. Formally, let $N$ denote the number of buses and $h_c$ the embedding dimension. The positional encoder is defined as a matrix $\mathbf{E} \in \mathbb{R}^{N \times h_c}$, where each row corresponds to a learned embedding associated with a bus index.}

\ra{The positional embeddings are added to the projected node feature vectors before graph convolution. This operation preserves node identity after spectral smoothing, mitigates over-smoothing effects in subsequent ARMAConv layers, and enhances the contrast between attacked and healthy regions of the grid. By introducing node-identity-aware positional information at an early stage, the positional encoder also facilitates effective global dependency modeling in the Transformer encoder.}

\subsection{ARMAConv}

\begin{figure}[ht]
\vspace{-.5cm}
    \centering
 \subfloat[\ra{One-hop aggregation from neighboring nodes in a grid-structured graph.}]{
    \includegraphics[clip, width=0.22\textwidth]{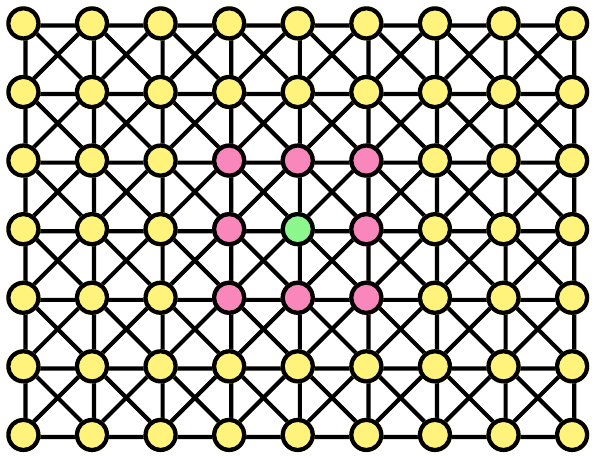}
    \label{fig:euclidean}
}
\hspace{0.01\textwidth}
\subfloat[\ra{One-hop aggregation from neighboring nodes in a non-Euclidean graph.}]{
    \includegraphics[clip, width=0.22\textwidth]{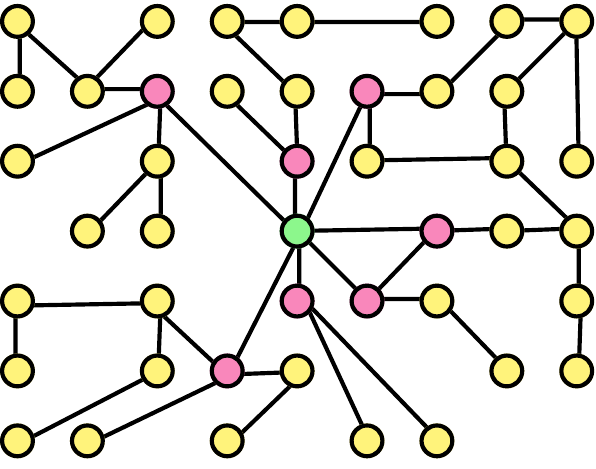}
    \label{fig:non-euclidean}
}
    \caption{\ra{Comparison of node information aggregation in grid-structured and non-Euclidean graphs. Graph-based convolution aggregates information from neighboring nodes in irregular topologies.}}
    \label{fig:two_diagrams}
\end{figure}

\ra{Graph convolutional operators are essential for processing power system data, as electrical networks exhibit non-Euclidean topologies that do not admit a fixed spatial ordering. As illustrated in Fig.~\ref{fig:two_diagrams}, conventional convolution assumes a regular grid structure and therefore does not directly extend to power grids, where each bus has a variable number of neighbors and irregular connectivity. Graph-based convolutional methods address this limitation by aggregating information from neighboring nodes according to the network topology, enabling localized feature extraction on irregular graphs.}

\begin{figure}
    \centering
    \includegraphics[width=0.45\textwidth]{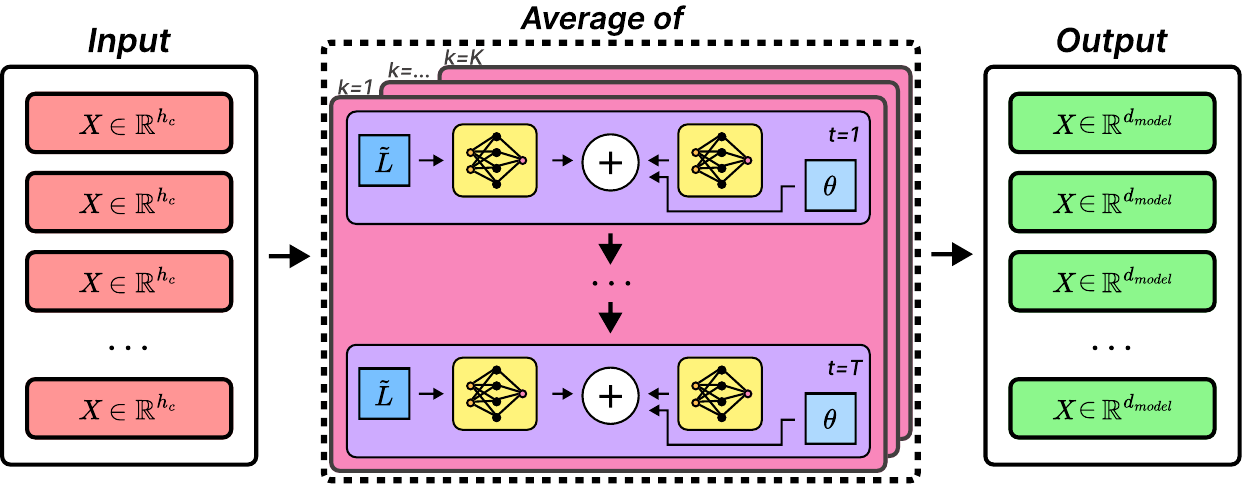}
    \caption{\ra{ARMAConv Architecture. The input consists of $N$ node feature vectors $X \in \mathbb{R}^{h_c}$, where $N$ is the number of buses in the power grid, and the output is a set of $N$ vectors with the same dimensionality. The dashed block illustrates Eq.~\ref{eq:arma-average}, which computes the average over $K$ parallel stacks. Each stack follows Eq.~\ref{eq:arma} and comprises $T$ recursive propagation steps, where each step combines the input processed by a neural network, a bias term $\theta$, and the modified Laplacian operator $\tilde{L}$ applied to a separate neural network. The output of each stack is propagated through the architecture, with the initial input passed recursively to the $T$-th step.}}
    \label{fig:armaconv}
\end{figure}

\ra{The ARMAConv graph filtering mechanism is employed in this work to extract topology-aware node representations while maintaining robustness to noise and preserving localization sensitivity. ARMAConv is particularly well suited for FDIA detection and localization in power grids, where system measurements typically vary smoothly across the network under normal operation but exhibit abrupt, localized deviations under malicious data injections~\cite{boyaci2021joint}. The overall ARMAConv architecture and its recursive message-passing mechanism are illustrated in Fig.~\ref{fig:armaconv}.}

\ra{To motivate the graph-domain formulation, we briefly recall the classical ARMA model from time-series analysis, which provides intuition for the spectral behavior of ARMA-based graph filters. An auto-regressive (AR) model of order $p$ computes a value $x_t$ at time $t$ as a weighted sum of past values:
\begin{equation}
x_t = \sum_{i=1}^{p} \phi_i x_{t-i} + \epsilon_t,
\end{equation}
where $\phi_i$ denotes the autoregressive coefficients and $\epsilon_t$ is a noise term. A moving average (MA) model of order $q$ expresses $x_t$ as a weighted sum of past noise terms:
\begin{equation}
x_t = \epsilon_t + \sum_{i=1}^{q} \theta_i \epsilon_{t-i},
\end{equation}
where $\theta_i$ are the moving average coefficients. Combining both components yields the ARMA$(p,q)$ model:
\begin{equation}
x_t = \sum_{i=1}^{p} \phi_i x_{t-i} + \epsilon_t + \sum_{j=1}^{q} \theta_j \epsilon_{t-j}.
\end{equation}}

\ra{While classical ARMA models operate on temporal sequences, they lack awareness of graph topology and may exhibit instability due to variance amplification. To extend ARMA filtering to graph-structured data, a graph shift operator is introduced via the normalized graph Laplacian
\begin{equation}
\mathbf{L}_{\text{norm}} = \mathbf{I} - \mathbf{D}^{-1/2} \mathbf{A} \mathbf{D}^{-1/2},
\end{equation}
where $\mathbf{A}$ is the adjacency matrix and $\mathbf{D}$ is the degree matrix. Using this operator, the graph ARMA filter defines spectral smoothing and diffusion across the network as
\begin{equation}
\mathbf{x}_{\mathbf{L}} =
\left(\mathbf{I} + \sum_{i=1}^{p} \phi_i \mathbf{L}^i \right)^{-1}
\left(\sum_{j=0}^{q} \epsilon_j \mathbf{L}^j \right),
\end{equation}
where higher powers of $\mathbf{L}$ correspond to information propagation across multi-hop neighborhoods.}

\ra{Direct computation of the inverse operator in the above expression is computationally expensive. The ARMAConv architecture proposed in \cite{bianchi2021graph} circumvents this limitation by iteratively approximating the inverse through message passing. Specifically, the recursive propagation rule is given by
\begin{equation}
\mathbf{X}^{(t+1)}_k = \sigma \left( \tilde{\mathbf{L}} \mathbf{X}^{(t)}_k \mathbf{W}
+ \mathbf{X}^{(0)} \mathbf{V} \right),
\label{eq:arma}
\end{equation}
where $\tilde{\mathbf{L}} = \mathbf{D}^{-1/2} \mathbf{A} \mathbf{D}^{-1/2}$, $k$ indexes parallel stacks, $t$ denotes the recursion depth, and $\mathbf{W}$ and $\mathbf{V}$ are learnable weight matrices. The final output is obtained by averaging the outputs of $K$ parallel stacks after $T$ recursive steps:
\begin{equation}
\mathbf{X}^{(T)} = \frac{1}{K} \sum_{k=1}^{K} \mathbf{X}^{(T)}_k.
\label{eq:arma-average}
\end{equation}}

\ra{The parallel-stack structure, depicted in Fig.~\ref{fig:armaconv}, enables ARMAConv to capture a broad range of spectral responses while improving robustness through ensemble averaging. Compared to polynomial-based graph convolutions such as Chebyshev graph convolutional networks (CGCNs), ARMAConv mitigates over-smoothing and noise amplification at higher filter orders \cite{cheng2024graph}. As a result, node-level identity is preserved more effectively, which is critical for accurate FDIA localization.}

\ra{In this work, layer normalization is applied to every other ARMAConv layer to further stabilize training and lower the computational cost by controlling feature magnitude growth. By combining localized smoothing with controlled depth and spectral diversity, ARMAConv provides a robust foundation for extracting discriminative, topology-aware representations suitable for detecting and localizing stealthy FDIAs in power grids.
}

\subsection{Encoder-Only Transformer}

\ra{Transformer architectures were originally introduced for sequence-to-sequence conversion; however, the encoder component has demonstrated strong representational capability even when used independently \cite{ding2023deot,sarrouti2022comparing}. In this work, an Encoder-Only Transformer is employed to model global dependencies among buses in a power grid, where the objective is to analyze a static snapshot of system measurements rather than generate sequential outputs. This design choice significantly reduces architectural complexity while retaining the ability to capture long-range, input-dependent interactions relevant to FDIA detection and localization.}

\begin{figure}[t]
    \centering
    \includegraphics[width=0.45\textwidth]{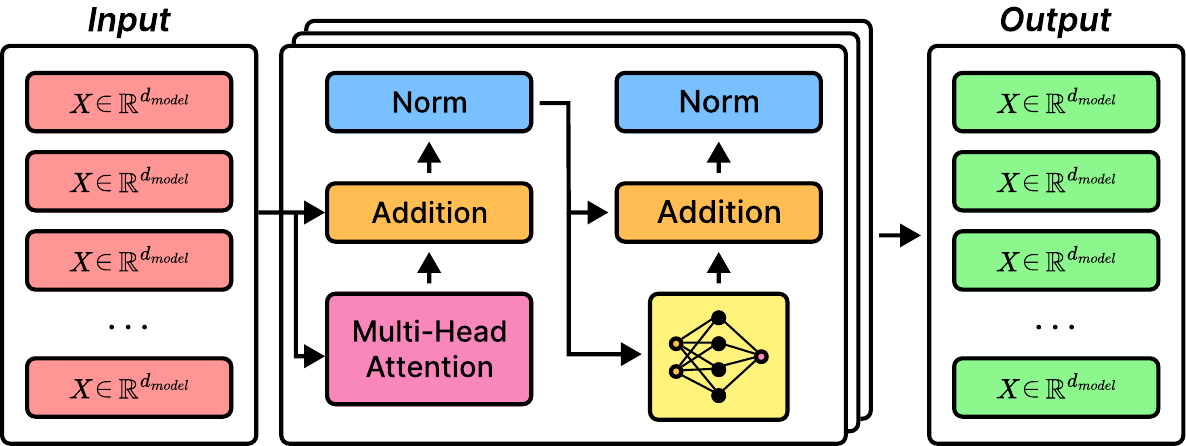}
    \caption{\ra{Encoder-Only Transformer architecture. Both the input and output consist of $N$ node embeddings $\mathbf{X} \in \mathbb{R}^{d_{model}}$. Each encoder layer comprises a multi-head self-attention sub-layer followed by a position-wise feed-forward network, with residual connections and normalization applied at each stage.}}
    \label{fig:encoder}
    \vspace{-.5cm}
\end{figure}

\ra{The encoder consists of a stack of identical layers, each taking the output of the previous layer as input. The input to the encoder is a sequence of node embeddings, where each bus is represented by a vector $\mathbf{X} \in \mathbb{R}^{d_{model}}$. Residual connections and normalization are applied to stabilize training and mitigate vanishing gradient issues in deep architectures \cite{xie2023residual}. As illustrated in Fig.~\ref{fig:encoder}, each encoder layer is composed of two sub-layers:
\begin{itemize}
    \item A multi-head self-attention sub-layer, which captures long-range dependencies among buses by allowing each node to attend to all others with a computational complexity of $O(n^2 d)$ per layer \cite{vaswani2017attention}.
    \item A position-wise feed-forward network (FFN) that applies a nonlinear transformation to enhance representational expressiveness:
    \begin{equation}
        FFN(x) = \sigma(x W_1 + b_1) W_2 + b_2,
        \label{eq:ffn}
    \end{equation}
    where $W_1$, $W_2$ are learnable weight matrices, $b_1$, $b_2$ are bias terms, and $\sigma(\cdot)$ denotes an activation function such as ReLU \cite{geva2020transformer}.
\end{itemize}
Residual connections are applied around each sub-layer, followed by normalization, enabling efficient gradient propagation and stable optimization.}

\begin{figure}[t]
    \centering
    \includegraphics[width=\columnwidth]{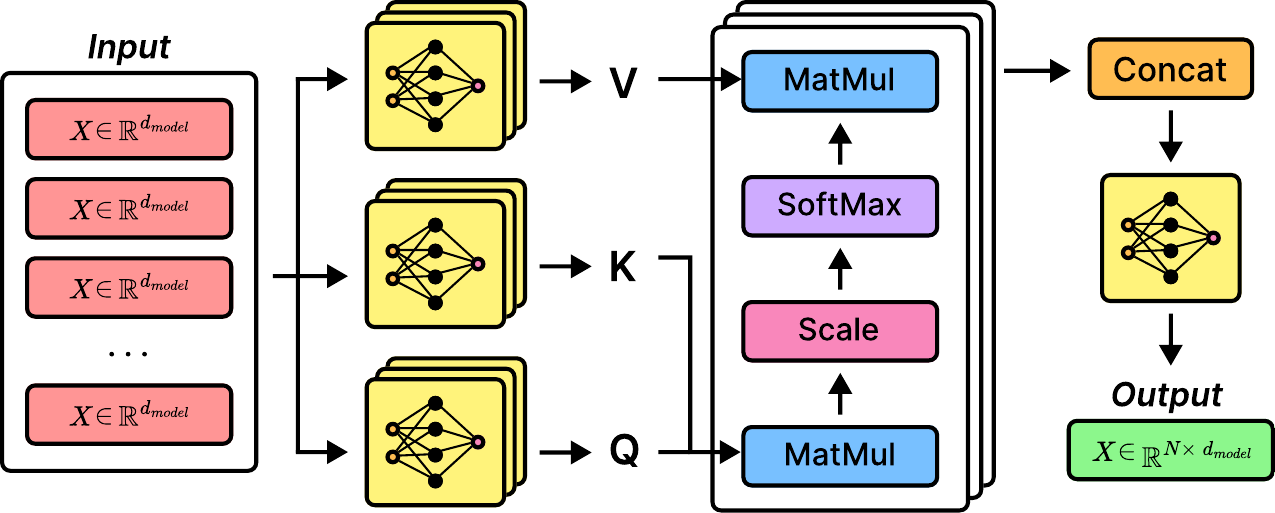}
    \caption{\ra{Scaled dot-product attention mechanism. Queries ($Q$), keys ($K$), and values ($V$) are obtained through linear projections of the input embeddings and combined via scaled dot-product attention to produce context-aware node representations.}}
    \label{fig:attentions}
    \vspace{-.5cm}
\end{figure}

\ra{The core mechanism enabling adaptive dependency modeling is multi-head self-attention, which has been shown to play a critical role in learning contextual representations \cite{voita2019analyzing}. Self-attention is based on the scaled dot-product formulation, where each input embedding is projected into \textit{queries} ($Q$), \textit{keys} ($K$), and \textit{values} ($V$) using learnable linear transformations. The attention operation computes a weighted aggregation of value vectors based on query–key similarity:
\begin{equation}
\text{Attention}(Q,K,V) = \text{softmax}\left(\frac{QK^{T}}{\sqrt{d_k}}\right)V,
\label{eq:attention}
\end{equation}
where $d_k$ denotes the dimensionality of the key vectors \cite{ghojogh2020attention}. The scaled dot-product attention mechanism is illustrated in Fig.~\ref{fig:attentions}.}
\ra{To increase modeling capacity and introduce parallel computation for scalability, the attention operation is performed across multiple parallel heads. Each head operates on a lower-dimensional projection, with dimensions given by
\begin{equation}
d_k = d_v = \frac{d_{model}}{h},
\label{eq:d-k}
\end{equation}
where $h$ is the number of attention heads. The outputs of individual heads are concatenated and linearly transformed to form the final multi-head attention output:
\begin{equation}
\!\!\!\!\text{MultiHead}(Q,K,V) = \mathcal{F}(head_1,\dots,head_h) W^O,
\label{eq:multihead-attention}
\end{equation}
with
\begin{equation}
head_i = \text{SelfAttention}(Q W_i^Q, K W_i^K, V W_i^V),
\label{eq:head-attentions}
\end{equation}
where $\mathcal{F}(\cdot)$ denotes concatenation and $W^O$ is a learnable projection matrix \cite{ding2022novel}.}

\ra{In the context of power systems, self-attention enables adaptive modeling of non-local electrical correlations and coordinated FDIA patterns that may span multiple regions of the grid. When combined with topology-aware representations from ARMAConv layers, the Encoder-Only Transformer provides a complementary mechanism for capturing global, input-dependent interactions without relying on topology-constrained message passing like standard GNN aggregation.}

\subsection{\rn{ARMAConv Encoder-Only Transformer}}

\ra{While positional encoding (PE) is effective in emphasizing node-level distinctions and mitigating feature homogenization, it is insufficient on its own for task-specific representation learning \cite{irani2025positional}. Similarly, ARMA-based graph convolutions excel at extracting topology-aware local features but may suffer from over-smoothing when stacked deeply, and experience diminished ability to capture global dependencies as network size increases\cite{bianchi2021graph,rusch2023survey}. Encoder-Only Transformer architectures, on the other hand, provide strong global modeling capability through self-attention but lack inherent awareness of graph structure and can be sensitive to noisy inputs \cite{zhang2024empirical,shen2023improving}. By integrating PE, ARMAConv, and an Encoder-Only Transformer into a unified architecture, the proposed model leverages the complementary strengths of each component while mitigating individual limitations.}

\begin{figure}[t]
    \centering
    \includegraphics[width=0.45\textwidth]{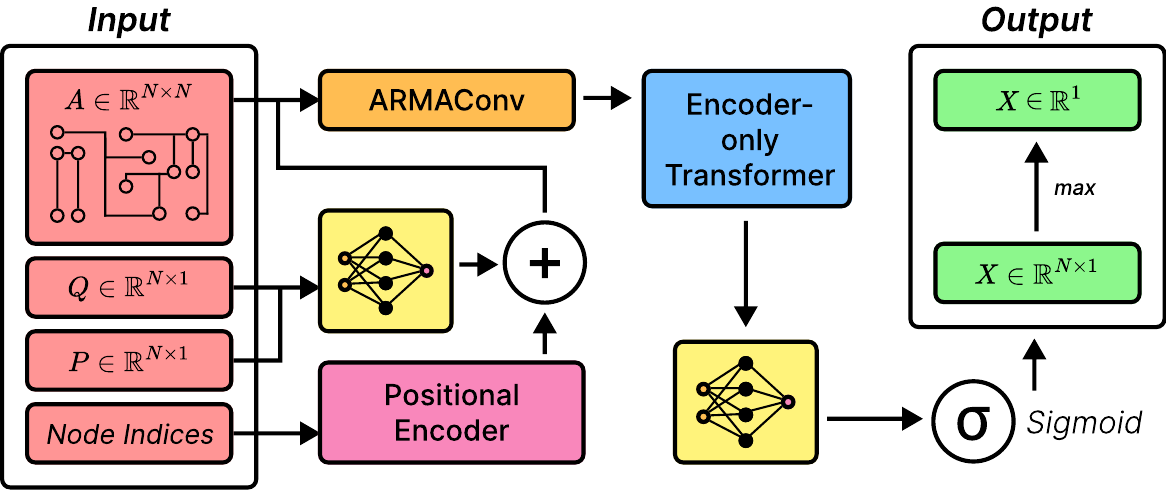}
    \caption{\rn{ARMAConv Encoder-Only Transformer (ACEOT) architecture. The model takes the weighted adjacency matrix, node-level power injections $(P,Q)$, and node indices as input, and outputs both node-level attack probabilities for localization and a graph-level probability for FDIA detection.}}
    \label{fig:aceot}
    \vspace{-.5cm}
\end{figure}

\ra{The overall processing pipeline is illustrated in Fig.~\ref{fig:aceot}. First, each bus is assigned a unique index, represented as a set of node identifiers $\mathbf{I} \in \mathbb{R}^{N}$, where $N$ denotes the total number of buses. These indices are passed to the positional encoder, which produces trainable embedding vectors that are added to the projected node feature representations to preserve node discriminability after spectral smoothing. In parallel, the raw power injection measurements of each bus, consisting of active and reactive power values $(P,Q)$, are linearly projected into the same embedding space.}

\ra{The resulting representations are summed to form the initial node feature matrix $\mathbf{X}^{(0)} \in \mathbb{R}^{N \times h_c}$ and passed, together with the weighted adjacency matrix $\mathbf{A}$, to the ARMAConv layers. ARMAConv performs topology-aware filtering to denoise the node features and aggregate localized neighborhood information while preserving node-level identity. The output of this stage is projected into $\mathbf{X}^{(1)} \in \mathbb{R}^{N \times d_{model}}$ and provided as input to the Encoder-Only Transformer.}

\ra{The Transformer encoder captures long-range, input-dependent interactions across all buses by adaptively weighting pairwise dependencies through self-attention. This mechanism enables the model to identify coordinated FDIA patterns that may span multiple regions of the grid and are not fully captured by localized spectral filtering alone. The final encoder output $\mathbf{X}^{(2)} \in \mathbb{R}^{N \times d_{model}}$ is mapped through a linear classification layer followed by a sigmoid activation to produce node-level attack probabilities $\mathbf{y} \in \mathbb{R}^{N \times 1}$.}

\ra{In addition to node-level localization, a graph-level detection output is derived by obtaining the maximum value among the node-level probabilities, indicating whether the power grid is under attack. This joint formulation supports both system-level awareness and targeted localization of compromised buses. By combining structure-aware filtering with adaptive global attention, the proposed ARMAConv Encoder-Only Transformer architecture alleviates over-smoothing, enhances robustness to noise, and improves FDIA detection and localization performance.}

\rn{\section{Experiments}}
\label{sec:experiments}

\ra{This section evaluates the proposed ACEOT framework for joint FDIA detection and localization. Experiments are conducted on standard IEEE test systems under realistic operating conditions derived from real-world load measurements and multiple FDIA scenarios. The evaluation assesses both detection performance and localization accuracy across a range of attack types and network sizes. Due to space limitations, detailed implementation procedures and hyperparameter configurations are reported in the Appendix.
}

\subsection{Dataset Generation}

\ra{Due to security and confidentiality concerns, real-world FDIA datasets are not publicly available, which limits the development of robust data-driven detection and localization methods. To address this challenge, we generate a synthetic dataset based on publicly available real-world load measurements and systematically modify them to simulate various false data injection attacks.}

\ra{Specifically, we employ the historical NYISO load profile for July 2021 to realistically reproduce temporal fluctuations in power system measurements. To increase both the resolution and size of the dataset, the original 5-minute interval data are interpolated to 1-minute intervals. Following Algorithm~1 in~\cite{boyaci2021graph}, we first generate a healthy (non-attacked) dataset for each time step}
\[
1 \leq t \leq 60 \times 24 \times 31
\]
for both IEEE-300 and IEEE-14 bus systems using the \texttt{Pandapower} Python library.

\ra{For each 1-minute interval, the network load and generation values are scaled by a stochastic factor}
\[
\mathcal{N}(1 + 0.1 S_t, \sigma_s),
\]
\ra{where $S_t$ represents the normalized load profile at time $t$ and $\sigma_s$ denotes the corresponding standard deviation. An AC power flow is then solved for the scaled network, after which 1\% relative uniform noise is added to the resulting measurements $(P_t, Q_t, V_t, \theta_t)$. The system state is subsequently estimated using the PSSE algorithm, and the estimated state is stored for later use either as a non-attacked sample or as a baseline for FDIA generation.}

\ra{The final dataset is constructed by introducing false data injections into the healthy samples. The number of data points corresponding to each FDIA type and dataset split is summarized in Table \ref{tab:dataset_split}. The resulting dataset consists of 34,560 samples. For each attacked instance, a time step is randomly selected without replacement from a uniformly shuffled list of all available time steps. A root bus is then randomly chosen, and a Breadth-First Search (BFS) is performed to identify all buses within a hop radius $r$, where $2 \le r \le 3$ for the IEEE-14 bus system and $6 \le r \le 8$ for the IEEE-300 bus system. Zero-injection buses and generator buses are excluded to avoid trivial attack patterns and to preserve dataset realism. FDIA perturbations are applied only to the buses identified by the BFS procedure, and the resulting instance is saved as an attacked sample.}

\ra{To emulate realistic cyberattack behavior, four FDIA types are considered. Two attacks are used during training, validation, and testing, while the remaining two are introduced exclusively during testing to assess the model’s generalization capabilities and ensure that we do not overfit the hyperparameters:}

\begin{itemize}
    \item \ra{\textbf{Refined constrained optimization-based FDIA ($A_o$)} adapted from~\cite{boyaci2021graph}. The original method is modified by replacing stochastic gradient descent with AdamW~\cite{loshchilov2017decoupled} to improve convergence. Additionally, the loss threshold $\tau_{\text{loss}}$ is determined using the Interquartile Range (IQR) method~\cite{whaley2005interquartile, vinutha2018detection}, computed from an empirical loss distribution obtained by applying the attack to all healthy samples.
    \item \textbf{Distribution-based FDIA ($A_d$)} following~\cite{yan2016detection, boyaci2021joint}, defined as
    \begin{equation}
        \mathbf{z}_a(t) \sim \mathcal{N}\left( \mu(\mathbf{z}_o), \sigma^2(\mathbf{z}_o) \right).
    \label{eq:distribution_fdia}
    \end{equation}
    \item \textbf{Data scaling attack ($A_s$)} used in~\cite{hasnat2020detection, jevtic2018physics}, defined as
    \begin{equation}
        \mathbf{z}_a(t) = \mathcal{U}(0.8, 1.2) \cdot \mathbf{z}_o(t).
        \label{eq:scaling_fdia}
    \end{equation}
    \item \textbf{Data replay attack ($A_r$)} following~\cite{shen2024detection}, defined as
    \begin{equation}
        \mathbf{z}_a(t) = \mathbf{z}_o(t - \tau),
        \label{eq:replay_fdia}
    \end{equation}
    where $\tau$ represents the replay delay and is set to $\tau = 4$ in this study.}
\end{itemize}

\ra{The final dataset is shuffled to eliminate seasonality effects and split into training, validation, and testing subsets using a $4{:}1{:}1$ ratio. Each subset contains an equal number of attacked and non-attacked samples. To reduce computational cost and mitigate numerical instability, all samples are standardized to zero mean and unit variance based on statistics computed from the training subset.}

\begin{table}[h!]
\centering
\caption{Number of data points in each subset}
\label{tab:dataset_split}
\resizebox{\columnwidth}{!}{
\begin{tabular}{| c | c c c c c | c |} 
 \hline
 Subset & $A_{o}$ & $A_{d}$ & $A_{s}$ & $A_{r}$ & Non-attacked & Total \\
 \hline\hline
 Training & 5760 & 5760 & 0 & 0 & 11520 & 23040 \\ 
 \hline
 Validation & 1440 & 1440 & 0 & 0 & 2880 & 5760 \\ 
 \hline
 Testing & 720 & 720 & 720 & 720 & 2880 & 5760 \\ 
 \hline
\end{tabular}
}
\end{table}

\subsection{Visualization of the Dataset Using t-SNE}

\ra{To confirm the validity of the dataset generated using the procedures described in the previous subsection, we visualize and analyze both the multidimensional input attributes and the corresponding targets. The high dimensionality of the data prevents direct visualization of feature correlations using conventional plotting techniques. At the same time, analyzing each feature independently would significantly limit our ability to observe global patterns and structural similarities across samples. To address these challenges, we employ t-distributed Stochastic Neighbor Embedding (t-SNE) to reduce the dimensionality of the data to a two-dimensional (2D) space and to uncover similarities between data points.}

\ra{The t-SNE algorithm minimizes the Kullback--Leibler (KL) divergence between probability distributions defined in the original high-dimensional space and the resulting low-dimensional embedding. In doing so, it maps each data point to a location in a 2D or 3D space such that points with similar feature representations remain close to one another~\cite{maaten2008visualizing}. This property allows t-SNE to effectively generate low-dimensional visualizations that contain well-defined clusters.}

\ra{In the context of FDIA detection, the presence and structure of clusters are particularly important. Specifically, clusters help verify that the modified healthy samples do not become obvious outliers as a result of data manipulation and therefore remain difficult to detect using simple threshold-based or statistical methods. This characteristic aligns with the stealthy nature of realistic false data injection attacks and motivates the use of t-SNE as a validation tool for the generated dataset.}

\begin{figure}[ht]
    \centering
    \subfloat[Node features]{
        \includegraphics[width=0.225\textwidth]{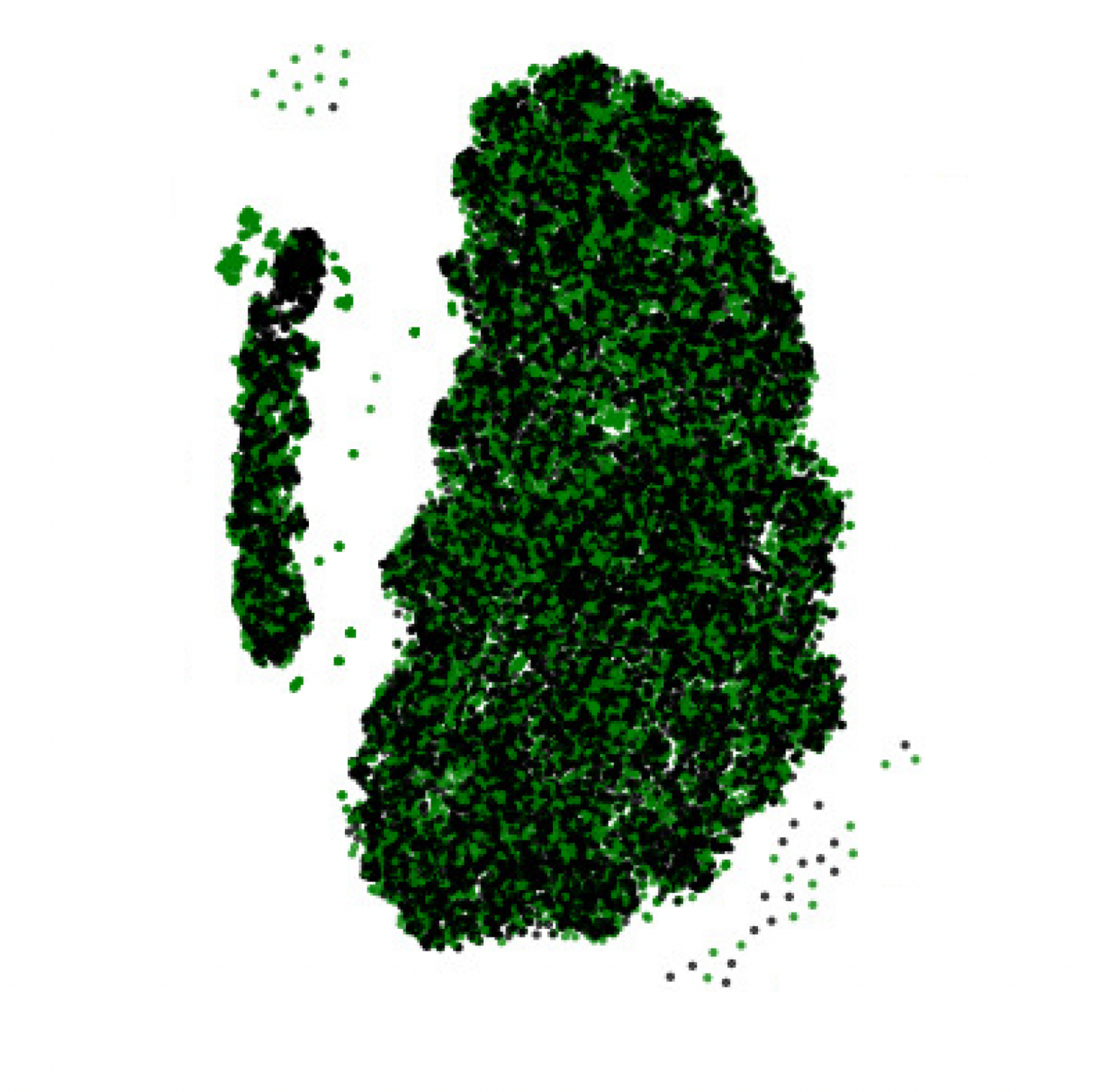}
        \label{fig:tsneX}
    }
    \hfill
    \subfloat[Node labels]{
        \includegraphics[width=0.225\textwidth]{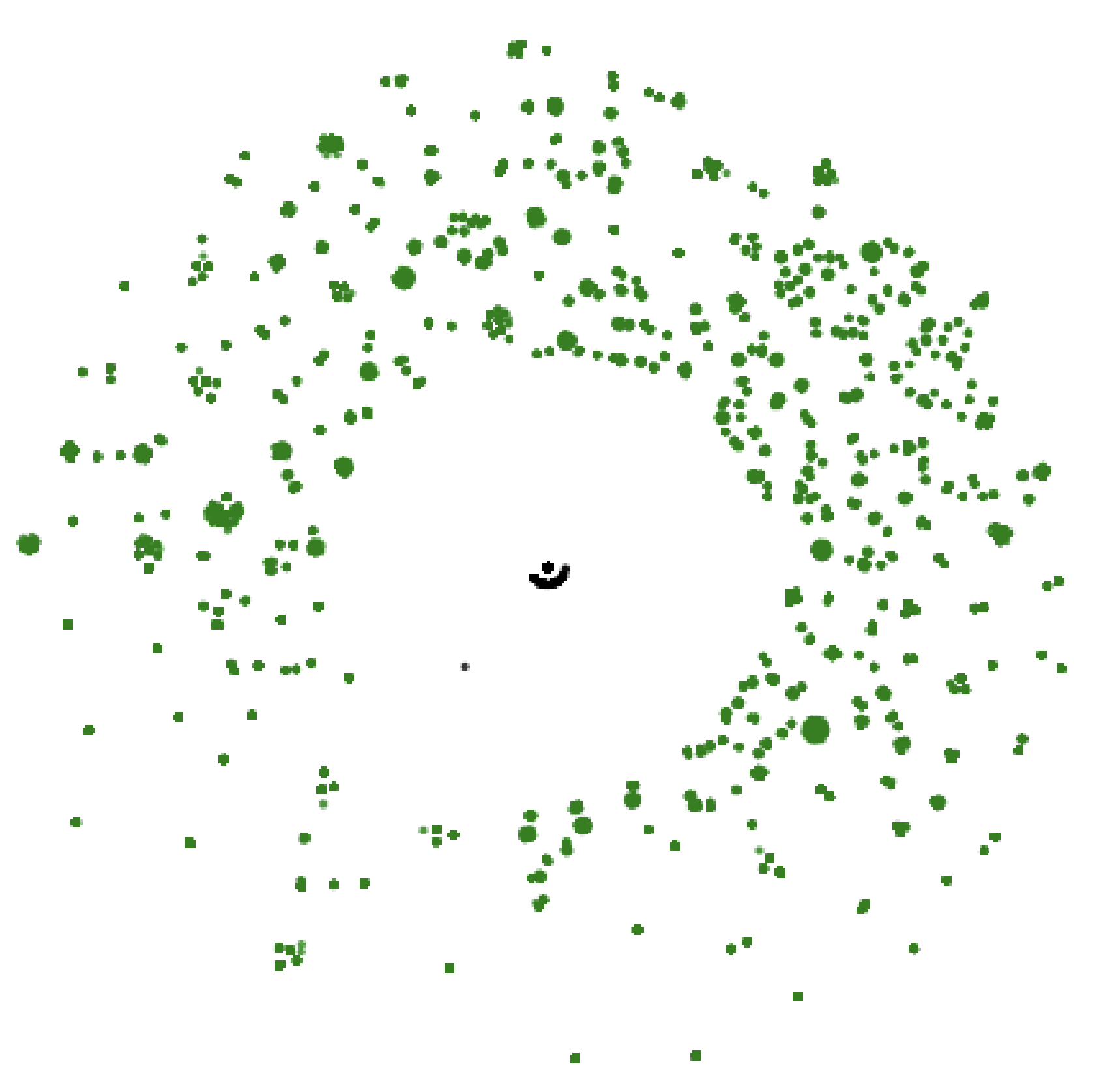}
        \label{fig:tsneY}
    }
    \caption{\ra{t-SNE visualization of IEEE-300 dataset samples.}}
    \label{fig:tsne_appendix}
\end{figure}

\ra{The 2D embeddings illustrated in Fig.~\ref{fig:tsneX} and Fig.~\ref{fig:tsneY} are obtained from the entire IEEE-300 bus system dataset consisting of 34,560 time steps. Fig.~\ref{fig:tsneX} presents the visualization of node-level input features $(P, Q) \in \mathbb{R}^{300 \times 2}$, where green points correspond to attacked samples at the graph level and black points represent healthy samples. The plot exhibits multiple distinct clusters, reflecting variations in the operating state of the power grid across different time periods. For example, even under normal (attack-free) conditions, the grid state can vary substantially between nighttime and peak-hour morning operations. Similar patterns arise when comparing weekday and weekend operating conditions.}

\ra{Importantly, each cluster in Fig.~\ref{fig:tsneX} contains both attacked and non-attacked samples, indicating a significant overlap between their feature representations. This observation is further reinforced by the absence of isolated single-colored clusters. Such overlap confirms that the injected attacks do not trivially separate from healthy data in feature space, thereby increasing the difficulty of detection and validating the realism of the generated FDIA scenarios.}

\ra{Fig.~\ref{fig:tsneY} visualizes the true node-level output labels $Y \in \mathbb{R}^{300}$. In this representation, non-attacked samples form a dense black core corresponding to identical zero-valued labels, while attacked samples appear as green points surrounding this core. This structure is expected, as healthy samples contain no attacked buses, whereas each attacked instance includes at least six connected buses with labels set to one. Together, the visualizations in Fig.~\ref{fig:tsneX} and Fig.~\ref{fig:tsneY} demonstrate that the dataset preserves realistic operating variability while embedding stealthy and spatially localized attack patterns.
}

\subsection{Model Training}

\ra{Each dataset sample corresponds to a network state consisting of 14 and 300 nodes for IEEE-14 and IEEE-300 bus networks, respectively, where each node is characterized by active ($P$) and reactive ($Q$) power measurements and an associated binary FDIA label. Voltage magnitude ($V$) and phase angle ($\theta$) are excluded due to their high correlation with $(P, Q)$ and limited empirical contribution to performance~\cite{liu2018data}.}

\ra{The training procedure is formulated as a multi-label supervised learning problem. The weighted adjacency matrix and node feature matrix serve as inputs, while node-level binary labels constitute the training targets. The AdamW optimizer is employed with a 500-step learning-rate warmup, where each step is a batch, followed by a cosine decay schedule. Binary cross-entropy loss with logits is used as the training criterion to improve numerical stability.}
\ra{Training is conducted using mini-batches of size 256 for a maximum of 256 epochs. Early stopping is applied to prevent overfitting, terminating training when the validation loss fails to decrease by more than $10^{-4}$ for 16 consecutive epochs. Hyperparameters for all models are tuned using the Optuna framework with 200 trials per model to ensure fair comparison. 
The resulting optimal hyperparameters are reported in Appendix~\ifarxiv the appendix\else \cite[Appendix]{abdulin2026attention}\fi.}

\subsection{Evaluation Metrics}

Performance is evaluated using Detection Rate (DR), False Alarm Rate (FA), and F1 score, defined as
\begin{align}
    \text{DR} &= \frac{\text{TP}}{\text{TP} + \text{FN}}, \\
    \text{FA} &= \frac{\text{FP}}{\text{FP} + \text{TN}}, \\
    \text{F1} &= \frac{2\,\text{TP}}{2\,\text{TP} + \text{FP} + \text{FN}}.
\end{align}

\ra{Due to the multi-label nature of the localization task, we evaluate performance from both sample-wise and node-wise perspectives. Sample-wise evaluation aggregates metrics across all buses for each time step, while node-wise evaluation aggregates metrics across all time steps for each bus. For concise localization assessment, we report the percentage of samples and nodes achieving F1 scores greater than 95\% and less than 5\%.}

\section{Results and discussion}
\label{sec:res_disc}
%In order to properly evaluate our model, we implement several ML benchmarks suitable for the multi-label classification and commonly utilized in the FDIA-related studies (e.g., \cite{zu2024self, wang2024fdia, sen2025neural, yang2021deep, drvodelic2024graphrt}): MLP, CNN, LSTM, CGCN, ARMAConv. Public implementations of these models are available in open-source Python libraries, such as PyTorch Geometric \cite{fey2019fast}, as well as in the following GitHub repository that contains all the Python algorithms employed in this research: https://github.com/netiRussell/Identification-and-Localization-of-Cyber-Attacks.

\ra{To comprehensively evaluate the proposed ACEOT model, we compare its performance against several machine learning baselines commonly used in FDIA detection and localization studies, including MLP, CNN, LSTM, ChebConv, and ARMAConv~\cite{zu2024self, wang2024fdia, sen2025neural}. Public implementations of these models are available through open-source Python libraries such as PyTorch Geometric~\cite{fey2019fast}, and all models are implemented and evaluated using the same testing pipeline. For transparency and reproducibility, the complete set of algorithms used in this study is available in the following public repository: \url{https://github.com/netiRussell/Identification-and-Localization-of-Cyber-Attacks}.}

\ra{Table~\ref{table:detection_table} summarizes the graph-level FDIA detection performance on the testing subset of the dataset in terms of Detection Rate (DR), False Alarm Rate (FA), and F1 score. An effective detection model is expected to achieve a high DR and F1 score while maintaining a low FA. In the IEEE-300 bus system, the proposed ACEOT model achieves the highest F1 score among all evaluated methods, while in the IEEE-14 system, it performs within 0.04\% of the best-performing baseline, ChebConv. ACEOT exhibits the lowest FA among complex models in the IEEE-300 case and remains competitive with ARMAConv in the IEEE-14 case.}

\begin{table}[h!]
\centering
\caption{Detection results in DR, FA, and F1 percentages}
\label{table:detection_table}
\resizebox{\columnwidth}{!}{
\begin{tabular}{|c|c|c|c|c|c|c|}
\hline
\multirow{2}{*}{\textbf{Model}} 
  & \multicolumn{2}{c|}{\textbf{DR (\%)}}
  & \multicolumn{2}{c|}{\textbf{FA (\%)}}
  & \multicolumn{2}{c|}{\textbf{F1 (\%)}} \\
\cline{2-7}
 & \textbf{IEEE-14} & \textbf{IEEE-300}
 & \textbf{IEEE-14} & \textbf{IEEE-300}
 & \textbf{IEEE-14} & \textbf{IEEE-300} \\
\hline\hline
 MLP & 27.92 & 24.72 &     0.66 & \textbf{0.10} &   43.42 & 39.61 \\ 
 \hline
 CNN & 83.26 & 64.34 &      2.08 & 0.62 &     89.85 & 78.00 \\
 \hline
 LSTM & \textbf{87.43} & 80.03 &     7.67 & 4.55 &      81.29 & 86.72 \\
 \hline
 ChebConv & 87.12 & 83.19 &  1.32 & 0.90 &     \textbf{92.46} & 90.38 \\
 \hline
 ARMAConv & 85.52 & \textbf{84.13} &     \textbf{0.56} & 0.90 &     91.92 & 90.94 \\
 \hline
 ACEOT & 83.58 & 83.26 &    1.28 & 0.42 &      92.42 & \textbf{92.60} \\ 
 \hline
\end{tabular}
}
\end{table}

%The Table \ref{table:detection_table}, as well as Figs. \ref{fig:boxplotSW} and \ref{fig:boxplotNW} contain the results obtained from the same testing algorithm applied to the "Testing" subset of the data.

\ra{Although ACEOT does not consistently achieve the highest DR, its performance reflects a deliberate tradeoff favoring precision over aggressive classification. This behavior can be interpreted as a ``cautious but accurate'' detection strategy, which prioritizes minimizing false alarms while preserving strong localization performance. Such characteristics are particularly desirable in power system operations, where excessive false alarms can undermine operator trust and lead to unnecessary actions.}

\begin{figure}
    \centering
    \includegraphics[width=0.45\textwidth]{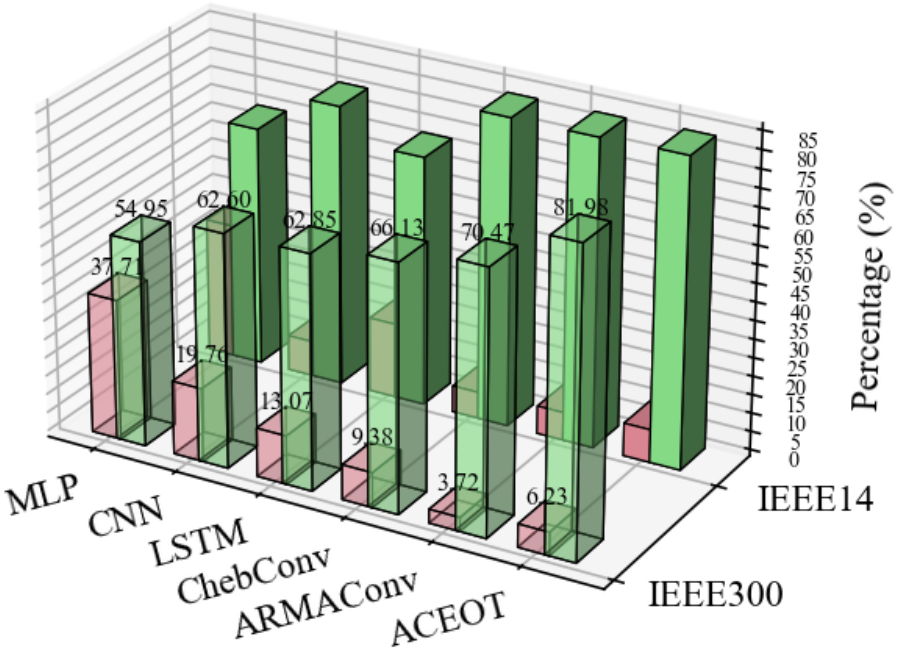}
    \caption{Sample-Wise evaluation of localization performance. Bars represent the percentage of buses that have an F1 score $\ge$ 95\% and $\le$ 5\%, shown in green and red, respectively.}
    \label{fig:boxplotSW}
\end{figure}

\ra{The sample-wise (SW) localization results are presented in Fig.~\ref{fig:boxplotSW}. Across both IEEE test systems, graph-based models that leverage Laplacian-based spectral filtering, namely ChebConv, ARMAConv, and ACEOT, significantly outperform non-graph baselines. For the IEEE-14 bus system, ChebConv, ARMAConv, and ACEOT achieve 82.76\%, 82.74\%, and 83.19\% of samples with F1 scores greater than or equal to 95\%, respectively. In the IEEE-300 bus system, the performance gap becomes more pronounced: ACEOT outperforms the next-best model, ARMAConv, by 11.51\% in the proportion of high-F1 samples.
}

\ra{This improvement provides quantitative evidence of the benefit introduced by the self-attention mechanism in ACEOT. Specifically, while ARMAConv relies solely on localized spectral message passing, ACEOT augments these representations with global, context-dependent interactions learned through attention. In large-scale networks such as IEEE-300, this additional modeling capacity enables ACEOT to better capture long-range dependencies between spatially distant but operationally correlated buses, resulting in a substantially higher proportion of accurately localized attack instances.
}

\ra{The positive correlation between network size and the performance advantage of ACEOT further supports this interpretation. As the number of nodes increases, purely spectral message-passing layers require a greater number of hops to propagate information across the graph, which can lead to over-smoothing and diminished sensitivity to localized anomalies. In contrast, the attention mechanism allows ACEOT to directly model interactions across the entire network, mitigating these limitations. Conversely, in smaller networks such as IEEE-14, fewer hops are sufficient to capture global structure, which reduces the relative advantage of attention-based modeling.
}

\ra{It should be noted that as models learn to generalize across diverse FDIA patterns, they may exhibit sensitivity to atypical or challenging scenarios, such as overlapping attacks, specific attack radii, or extreme load conditions. While ACEOT achieves higher overall localization accuracy, it also yields a slightly higher proportion of samples with F1 scores below 5\% compared to ARMAConv (a difference of 2.51\%). This behavior reflects the model’s specialization toward learning rich contextual representations, which improves average performance but may underperform in rare or anomalous cases. In contrast, ARMAConv, which does not employ attention, learns more uniform representations that result in lower peak performance but fewer failure cases.
}

\begin{figure}
    \centering
    \includegraphics[width=0.45\textwidth]{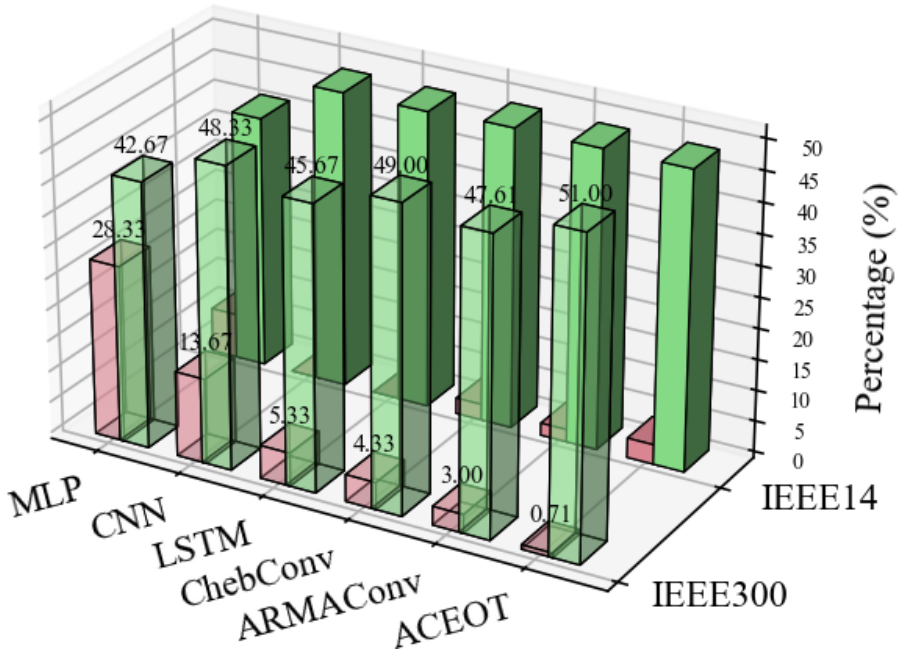}
    \caption{Node-Wise evaluation of localization performance. Bars represent the percentage of buses that have an F1 score $\ge$ 95\% and $\le$ 5\%, shown in green and red, respectively.  }
    \label{fig:boxplotNW}
\end{figure}

\ra{The node-wise (NW) localization results are shown in Fig.~\ref{fig:boxplotNW}. Compared to SW evaluation, NW scores are consistently lower across all models due to class imbalance, as many buses are rarely attacked and therefore have limited positive labels. Nevertheless, NW evaluation provides a meaningful measure of how reliably individual buses can be classified across time. In both IEEE test systems, ACEOT achieves the highest NW localization performance. For the IEEE-14 case, CNN, LSTM, ChebConv, and ARMAConv reach 47.86\%, 47.86\%, 48.30\%, and 48.16\% of buses with F1 scores exceeding 95\%, respectively, while ACEOT attains 49.09\%. Similar trends are observed in the IEEE-300 system, where the benefits of attention-based modeling are more pronounced.
}

\ra{Finally, we evaluate the computational efficiency of all models by measuring the average inference time required to process a single network state. For the IEEE-14 bus system, MLP, CNN, LSTM, ChebConv, ARMAConv, and ACEOT require 10.47~ms, 10.83~ms, 8.10~ms, 25.83~ms, 15.55~ms, and 15.85~ms, respectively. For the IEEE-300 bus system, the corresponding inference times are 10.61~ms, 11.01~ms, 9.84~ms, 27.76~ms, 20.49~ms, and 20.96~ms. As expected, LSTM achieves the lowest inference time due to its relatively small number of parameters and lack of graph-structured computations. Notably, integrating the Transformer Encoder with ARMAConv does not introduce a significant computational burden, as attention heads are processed in parallel on the GPU. As a result, ACEOT achieves inference times comparable to ARMAConv while substantially improving localization performance in large-scale networks.}

\section{Conclusion}
\label{sec:Conclusion}
%This paper introduced the ACEOT model, a novel algorithm for joint FDIA detection and localization. The resulting model can be described as a combination of positional encoding, ARMAConv, and Encoder-Only Transformer components that together stress the corrupted nodes of a power system based on the physics of a network. The model manages to capture both local and global spatial dependencies of the grid layout and its measurements by utilizing a self-attention mechanism with encoded graph convolutions. The proposed method outperforms the state-of-the-art ARMAConv and ChebConv models. These results are further supported by the fact that the simulation was conducted on a standard IEEE-300 and IEEE-14 bus networks using the publicly available NYISO data, confirming that our solution can effectively manipulate real-world-like global and long-range spatial information of different sizes to help prevent FDIAs online. 

\ra{This paper presented ACEOT, a novel framework for joint false data injection attack (FDIA) detection and localization in power systems. The proposed model integrates positional encoding, convolutional auto-regressive moving average (ARMAConv) graph filtering, and an Encoder-Only Transformer architecture to identify corrupted measurements in a manner consistent with the underlying physics and topology of the power grid. By combining topology-aware spectral filtering with a self-attention mechanism, ACEOT effectively captures both local and long-range spatial dependencies present in grid measurements.
Extensive evaluations conducted on standard IEEE-14 and IEEE-300 bus systems using realistic operating conditions derived from publicly available NYISO load data demonstrate that ACEOT consistently outperforms state-of-the-art ARMAConv and ChebConv-based methods in both detection and localization tasks. The results further indicate that the advantages of attention-based modeling become more pronounced as network size increases, highlighting the scalability of the proposed approach. Collectively, these findings confirm that ACEOT can reliably exploit global contextual information and long-range correlations in large-scale power systems, offering an effective and practical solution for online FDIA detection and localization.
}

\ifarxiv
\appendix
\section{Appendices}
\section{Hyperparameter Settings}
\label{app:hyperparameters}

\ra{This appendix reports the optimal hyperparameters obtained using the Optuna framework for all models considered in this study. Each model was tuned using 200 trials with identical search budgets to ensure fair comparison.}

\begin{table*}[htbp]
\centering
\caption{Model-specific hyperparameter tuning results}
\label{tab:model_hyperparameters}
\begin{tabular}{l l l c c}
\hline
\textbf{Model} & \textbf{Hyperparameter} & \textbf{Options} & \textbf{IEEE-14} & \textbf{IEEE-300} \\
\hline
MLP & Layers & \{2,3,4\} & 3 & 3 \\
 & Hidden units & \{32,64,128\} & 64 & 128 \\
CNN & Layers & \{2,3,4\} & 3 & 3 \\
 & Kernel size & \{2,3,4\} & 3 & 3 \\
 & Hidden units & \{32,64,128\} & 64 & 64 \\
LSTM & Layers & \{2,3,4\} & 3 & 4 \\
 & Hidden units & \{32,64,128\} & 32 & 64 \\
ChebConv & Layers & \{2,3,4\} & 3 & 3 \\
 & Chebyshev order & \{3,4,5,7\} & 3 & 4 \\
 & Hidden units & \{16,32,64,96,128\} & 64 & 96 \\
ARMAConv & Layers & \{2,3,4\} & 3 & 3 \\
 & Stacks & \{3,4,5\} & 2 & 3 \\
 & Iterations & \{4,5,6\} & 3 & 5 \\
 & Hidden units & \{32,64,96,128\} & 32 & 64 \\
ACEOT & ARMAConv layers & \{2,3,4\} & 3 & 3 \\
 & Transformer heads & \{2,4,8\} & 4 & 4 \\
 & Transformer layers & \{1,2,3\} & 1 & 2 \\
 & $d_{\text{model}}$ & \{64,128,256\} & 64 & 256 \\
 & FFN dimension & \{128,256,512\} & 128 & 512 \\
\hline
\end{tabular}
\end{table*}

\begin{table*}[htbp]
\centering
\caption{General hyperparameter tuning results}
\label{tab:general_hyperparameters}
\begin{tabular}{l c c c c c c}
\hline
\textbf{Model} & \multicolumn{2}{c}{\textbf{Learning rate}} & \multicolumn{2}{c}{\textbf{Dropout}} & \multicolumn{2}{c}{\textbf{Positive weight}} \\
 & IEEE-14 & IEEE-300 & IEEE-14 & IEEE-300 & IEEE-14 & IEEE-300 \\
\hline
MLP & $10^{-3}$ & $10^{-3}$ & 0.20 & 0.30 & 1.0 & 1.0 \\
CNN & $10^{-3}$ & $10^{-3}$ & 0.15 & 0.15 & 2.0 & 2.0 \\
LSTM & $1.5\!\times\!10^{-2}$ & $2.25\!\times\!10^{-2}$ & 0.25 & 0.10 & 5.0 & 8.4 \\
ChebConv & $10^{-2}$ & $10^{-2}$ & 0.10 & 0.10 & 3.0 & 3.0 \\
ARMAConv & $2\!\times\!10^{-3}$ & $10^{-3}$ & 0.10 & 0.20 & 3.0 & 1.5 \\
ACEOT & $10^{-3}$ & $10^{-3}$ & 0.25 & 0.20 & 8.5 & 9.0 \\
\hline
\end{tabular}
\end{table*}

\fi

\bibliographystyle{IEEEtran}
\bibliography{ref}
\end{document}